\documentclass[review]{elsarticle}
\graphicspath{ {./figures/} }
\usepackage{hyperref}
\usepackage{float}
\usepackage{verbatim} 
\usepackage{apalike}
\usepackage{color}
\usepackage{amsmath,amsthm,amssymb,amsfonts}
\usepackage{booktabs}

\usepackage{graphicx}
\usepackage{multirow}
\usepackage{ulem}
\usepackage{tcolorbox}
\newcommand{\smallcolor}[1]{
	{\underline{#1}}
}

\newcommand{\bigcolor}[1]{
	{\underline{#1}}
}

\restylefloat{figure}
\restylefloat{table}

\journal{Expert Systems with Applications}

\biboptions{authoryear}

\begin{document}
	\begin{frontmatter}

		\begin{titlepage}
			\begin{center}
				\vspace*{1cm}
				
				\textbf{ \large A Novel Prompt-tuning Method:\\Incorporating Scenario-specific Concepts into a Verbalizer}
				
				\vspace{1.5cm}
				
				Yong Ma$^{a,c}$ (znsoft@163.com), Senlin Luo$^a$ (luosenlin2019@126.com), Yu-Ming Shang$^b$ (shangym@bupt.edu.cn), Zhengjun Li$^{a}$ (zjli@bit.edu.cn), Yong Liu$^c$ (liuyong03@qianxin.com) \\
				
				\hspace{10pt}
				
				\begin{flushleft}
					\small  
					$^a$ School of Information and Electronics, Beijing Institute of Technology, Haidian District, Beijing, China \\
					$^b$ School of Cyberspace Security, Beijing University of Posts and Telecommunications, Haidian District, Beijing, China \\
					$^c$ Qi-AnXin Technology Group, QAX Security Center, Xicheng District, Beijing, China

				\end{flushleft}        
			\end{center}
		\end{titlepage}
\begin{abstract}
The verbalizer, which serves to map label words to class labels, is an essential component of prompt-tuning. In this paper, we present a novel approach to constructing verbalizers. While existing methods for verbalizer construction mainly rely on augmenting and refining sets of synonyms or related words based on class names, this paradigm suffers from a narrow perspective and lack of abstraction, resulting in limited coverage and high bias in the label-word space. To address this issue, we propose a label-word construction process that incorporates scenario-specific concepts. Specifically, we extract rich concepts from task-specific scenarios as label-word candidates and then develop a novel cascade calibration module to refine the candidates into a set of label words for each class. We evaluate the effectiveness of our proposed approach through extensive experiments on {five} widely used datasets for zero-shot text classification. The results demonstrate that our method outperforms existing methods and achieves state-of-the-art results. The source code of ISCV is openly accessible via the following URL: https://github.com/BIT-ENGD/ISCV\_ARXIV.

\end{abstract}

		
\end{frontmatter}
	
	\section{Introduction}
	\label{introduction}

	Pre-trained language models (PLMs) \citep{hanPreTrainedModelsPresent2021a} have emerged as a fundamental and powerful instrument for natural language processing (NLP). They are applied in a wide range of downstream NLP tasks, including text classification \citep{howardUniversalLanguageModel2018a}, question answering \citep{petroniLanguageModelsKnowledge2019a}, and machine translation \citep{radfordLanguageModelsAre2019}. Notably, PLMs store a vast amount of linguistic knowledge \citep{petroniLanguageModelsKnowledge2019a}, which has attracted the attention of numerous researchers seeking to explore and leverage the rich knowledge contained in PLMs. In this regard, the prompt-tuning approach is a representative means of achieving this goal.
	
\begin{figure}[t]
	\centering
	\includegraphics[width=1\columnwidth]{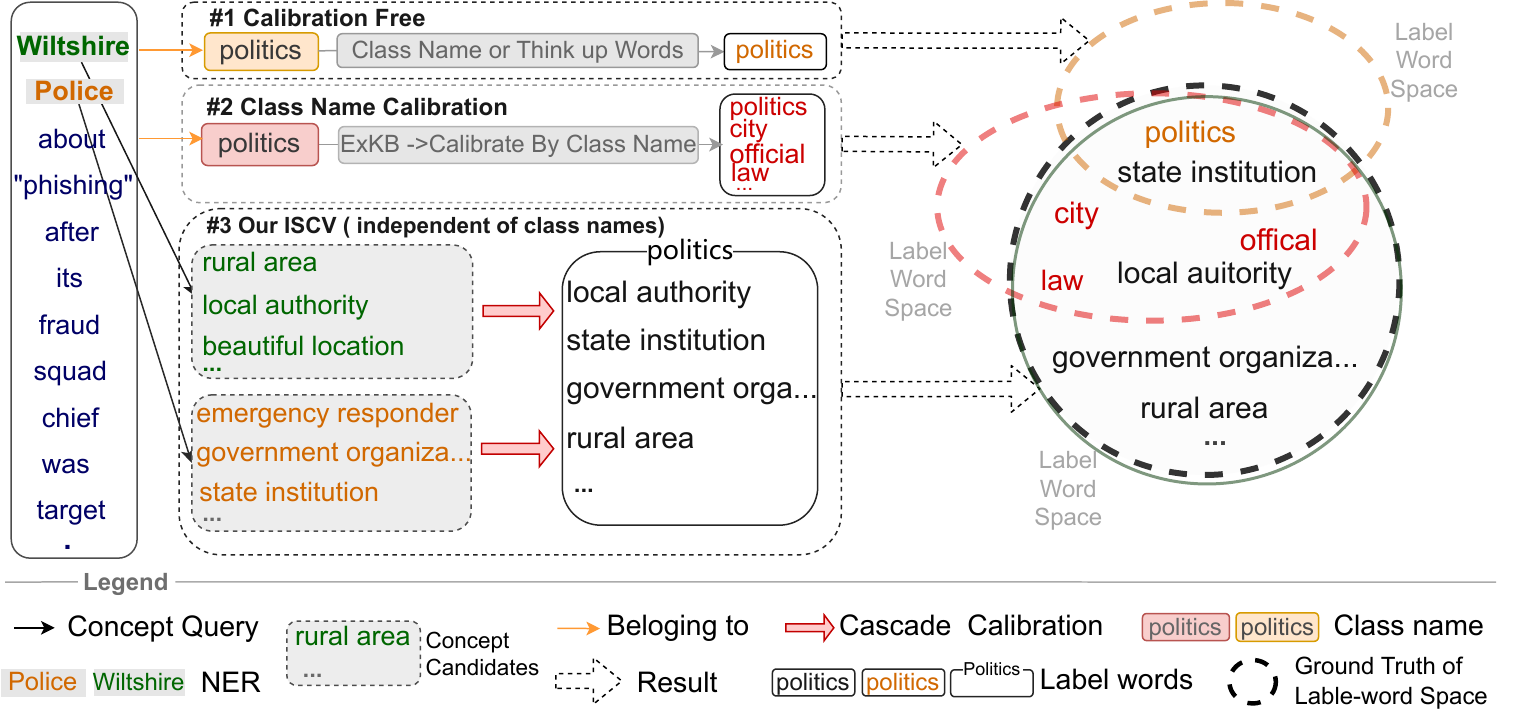} 
	\caption{
		{Illustration of the ISCV motivation.} ``ExKB" means retrieving words from an external knowledge base. Expansion from multiple perspectives and higher-level abstractions, as well as task-specific refinement, enables large coverage and low bias in a label-word space.
	}
	\label{motivation}
\end{figure}

A prompt-tuning approach consists of three main components: a PLM, a prompt template, and a verbalizer.
The PLM is used to predict the probability of each word in its vocabulary at the masked location. The prompt template is a piece of natural language text with masked locations to elicit knowledge from a PLM. 
A verbalizer serves as a projecting module between label words and class labels \citep{schickExploitingClozeQuestionsFewShot2021}, for example, ``company" $\mapsto$ ``BUSINESS".
Given a sentence ``Apple is a giant electronic company.", to classify it into the class ``BUSINESS",  it needs to wrap the sentence into a prompt template as ``It is a [MASK] news, Apple is a giant electronic company", then feed it into the PLM. 
If the probability of predicting the label word ``company" filled in ``[MASK]" location is greater than a threshold, assign the sentence a class label ``BUSINESS". Therefore, a verbalizer plays a crucial role in the prompt-tuning method \citep{gaoMakingPretrainedLanguage2021a}.

Existing verbalizer construction methods can be broadly categorized into two groups: calibration-free methods and class-name calibration methods. As \#1 in Figure \ref{motivation}, the former employs a list of  hand-crafted synonyms or related words \citep{schickItNotJust2021a} even directly only the class names as the label words of a class without a calibration processing. As \#2 in Figure \ref{motivation}, the latter augments additional related words but refines them merely based on class names \citep{huKnowledgeablePrompttuningIncorporating2022}. 
However, a calibration-free approach generates a few label words normally, which is limited by the number of label words and the designer's knowledge limitations, and thus has fewer perspectives in a label-word space. Meanwhile, the class name based calibration methods expand additional label words and refine them based on the class names, while class names are not the ideal expanding center in general, and some effective label words are not among the expanded words, which somewhat results in a deviation from the true center. Therefore, all the above methods encounter fewer perspectives and lower-level abstractions, resulting in narrow coverage and high bias in a label-word space.

Previous research suggests \citep{jiMicrosoftConceptGraph2019}  that humans understand an instance, especially an unfamiliar instance, by its basic concepts at an appropriate level. A concept is defined as Basic-level Categorization (BLC) by psychologists and linguists, which has higher-level abstraction than the instance \citep{eysenckFundamentalsCoginition2018}, thus having additional coverage.
Intuitively, label-word coverage can be improved if we can augment label words from multiple perspectives and higher-level abstractions by leveraging instance concepts.
Obviously, the above intuition may introduce noisy concepts that are not relevant to our tasks, leading to possible over-coverage, so we should refine the augmented label words.

Drawing inspiration from the above intuition, we present a new approach for prompt tuning called \textit{Incorporating Scenario-specific Concepts into Verbalizer construction} (ISCV). The term "scenario-specific" refers to the scenario that is associated with specific task information, which involves a particular dataset and prompt template, and is also referred to as task-specific in the following discussion.
Specifically, the verbalizer construction of ISCV method has two steps: concept mining and cascade calibration. 
(1) To mine rich scenario-specific concepts, a set of samples is randomly selected from a task-specific scenario and then associated concepts are obtained based on them as label-word candidates.
(2) To denoise irrelevant or ineffective label-word candidates caused by over-coverage, a novel cascade calibration approach is proposed, which consists of a PLM based calibration procedure and a category based calibration procedure. 
Therefore, the constructed verbalizer can be directly applied in a prompt-tuning pipeline, which can boost the performance of target tasks.
Extensive experimental results of zero-shot text classification tasks on {five widely used text }classification datasets demonstrate the effectiveness of our approach,  showing that ISCV achieves fresh state-of-the-art results.
Moreover, ISCV can also reduce the standard deviation of results, enabling better stability of the experimental results among different prompt templates.


The contribution of this paper can be summarized as:
\begin{itemize}
	\item We present a fresh perspective on incorporating scenario-specific concepts into verbalizer construction. The present work is the first to employ multiple perspectives and high-level abstractions to develop label words based on external conceptual knowledge and scenario-specific settings.
	\item We devise a novel cascade calibration module based on scenario information instead of relying on class names to refine the label word set for a verbalizer.
	
	  \item Fresh state-of-the-art results on {three topic classification and two sentiment analysis} datasets are reported, and extensive analysis is conducted to provide reasons behind the outstanding effectiveness of our approach.
\end{itemize}

The subsequent sections of this paper are organized as follows: Section 2 provides a review of related work, while Section 3 presents the details of our proposed method. We describe the experimental setup and provide related discussions in Sections 4 and 5, respectively. Finally, we present our conclusions in Section 6.

\section{Related Work} 
This paper presents a novel scenario-specific verbalizer-construction method for prompt-tuning. In this section, we provide a review of two groups of related research.

\subsection{Prompt-tuning}

Since the emergence of GPT-3 \citep{brownLanguageModelsAre2020}, prompt-tuning, also known as prompt-based learning, has garnered increasing attention \citep{liuPretrainPromptPredict2021a}.
Existing research focusing on prompt-tuning comprises three key components: a pre-trained language model (PLM), a prompt template, and a verbalizer.
Prior studies on PLMs have primarily focused on their architecture, including BERT \citep{devlinBERTPretrainingDeep2019a}, RoBERTa \citep{liuRoBERTaRobustlyOptimized2019a}, GPT-3, BART \citep{lewisBARTDenoisingSequencetoSequence2020}, T5 \citep{raffelExploringLimitsTransfer2020a}, and others. Research on prompt templates involves developing discrete templates \citep{jiangHowCanWe2020a,shinAutoPromptElicitingKnowledge2020a,petroniLanguageModelsKnowledge2019a,yuanBARTScoreEvaluatingGenerated2021} and continuous templates {\citep{liPrefixTuningOptimizingContinuous2021a,shinAutoPromptElicitingKnowledge2020a,zhangDifferentiablePromptMakes2021,xuMakingPretrainedLanguage2023}}.
Recent studies have focused on constructing continuous verbalizers {\citep{hambardzumyanWARPWordlevelAdversarial2021b,cuiPrototypicalVerbalizerPromptbased2022,xuMakingPretrainedLanguage2023}}, while others have concentrated on discrete verbalizers {\citep{huangFewShotFineGrainedEntity2022,huKnowledgeablePrompttuningIncorporating2022}}. For simplicity, we refer to discrete verbalizers as ``verbalizers" throughout this paper. The approach presented in this paper is dedicated to the construction of discrete verbalizers.

\subsection{Verbalizer Construction}
Existing verbalizer construction approaches can be broadly categorized into two groups: calibration-free methods and class-name calibration methods. Among the former, Yin et al. \citep{yinBenchmarkingZeroshotText2019a} manually design word lists related to class names as label word sets. Schick et al. \citep{schickExploitingClozeQuestionsFewShot2021} leverage the class names directly as label words, resulting in limited coverage due to the fewer label words. For the latter, Hu et al. \citep{huKnowledgeablePrompttuningIncorporating2022} expand related words from external knowledge bases as label words and refine them based only on class names.
However, the label words generated by the aforementioned methods are generally synonyms or related words of class names, which have fewer perspectives and lower-level abstractions. Moreover, these words have been calibrated based solely on class names or not at all, resulting in low coverage and high bias in the label-word space.

In contrast to existing methods, the proposed approach leverages scenario-specific concepts to construct a verbalizer, resulting in improved perspectives and abstraction levels of label words. Additionally, the approach refines label words based on scenario-specific information, enhancing coverage, and reducing bias in the label-word space.

\section{Method}

\begin{figure*}[t]
	\centering
	\includegraphics[width=1\columnwidth]{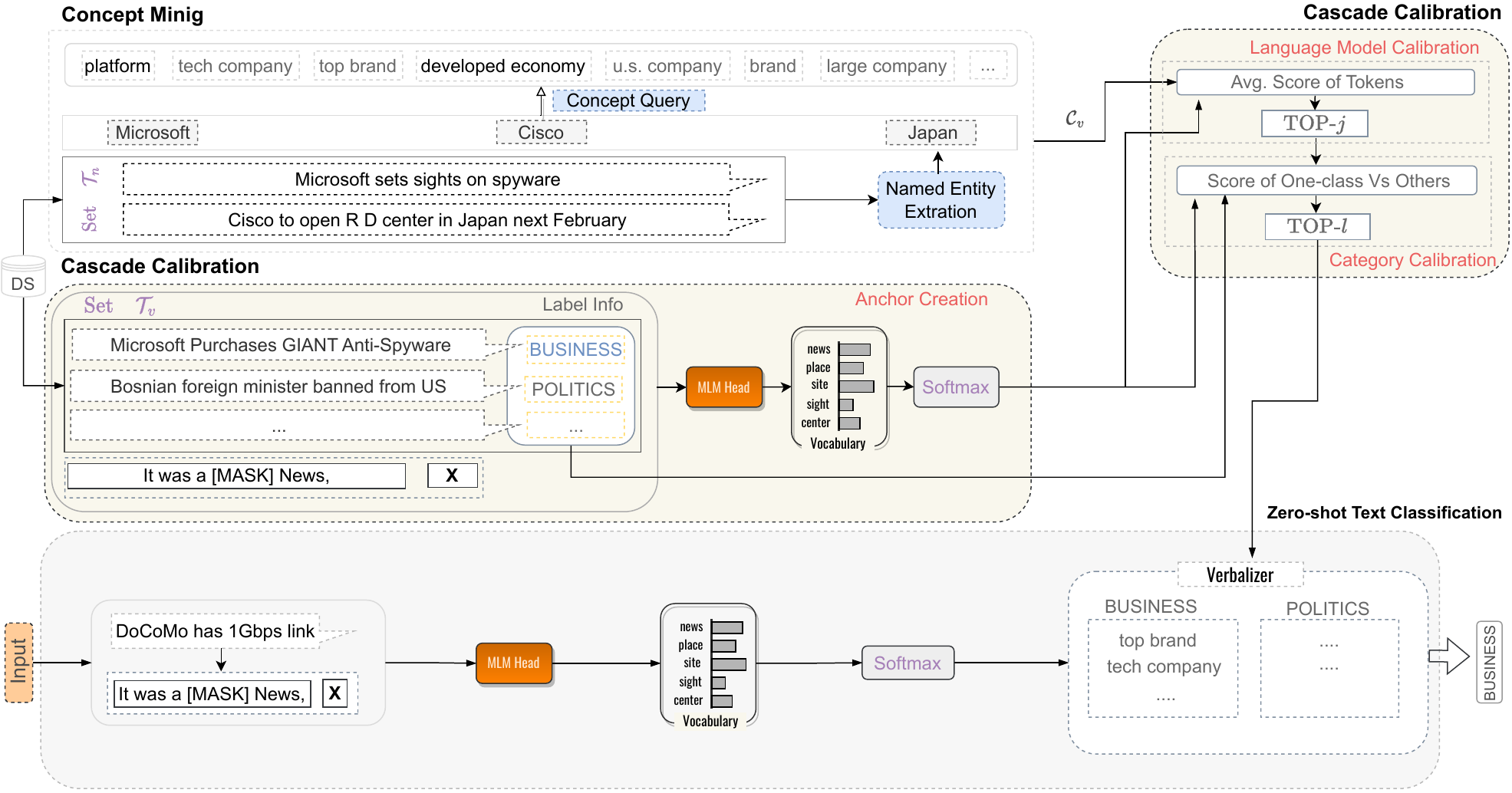} 
	\caption{
		The architecture of our ISCV including a verbalizer construction approach and a zero-shot learning pipeline. The verbalizer construction approach consists of a concept mining procedure and a cascade calibration procedure. The two light-beige segments constitute the cascade calibration procedure. The section with white color is the concept mining procedure. The bottom part is the zero-shot-based text classification pipeline in prompt-tuning. DS means a dataset.
	}
	\label{architecture}
\end{figure*}

An overview of our ISCV is depicted in Figure \ref{architecture}, which includes a verbalizer construction approach in the top half part, colored in light beige and white, and a zero-shot based text classification pipeline in the bottom half part.
In this paper, we focus on the verbalizer construction approach, which consists of two procedures: concept query and cascade calibration. A set of support samples, $\mathcal{T}_n$, are fed into the concept query procedure to obtain a set of concepts as candidates. At the same time, another set of samples with label information, $\mathcal{T}_v$, are fed into the anchor creation step to obtain task-specific normalization exponents as anchors. Both the candidate concepts and the anchors are fed into the language model calibration step and then run through the whole cascade calibration procedure to refine concepts as label-word sets for a verbalizer.
The constructed verbalizer can be seamlessly integrated into a prompt-tuning pipeline.
In the following, we detail the above procedure, where the verbalizer construction is our focus.

\subsection{Task Definition}

In a prompt-tuning based text classification task, a input text is wrapped into a prompt template as a natural language prompt fed into a PLM, thus the task is reformulated as a cloze-style task \citep{schickItNotJust2021a}.
Let  $\mathcal{M}$ be a pre-trained language model, $\mathbf{x} =\{w_0,w_1,\dotsm,w_n\}$ be an input text piece to be categorized into a class label $y\in \mathcal{Y}$, $\mathcal{Y}=\{y_0,y_1,\dotsm,y_k\}$. 
The input text $\mathbf{x}$ is wrapped into a prompt template as
\begin{equation*}
	\mathbf{x}_{\text{p}}=\text{\texttt{[CLS]} A \texttt{[MASK]} news : } \mathbf{x},
\end{equation*}
{where $\mathbf{x}_{\text{p}}$ is referred to as a prompt or contextual sample that is fed into the model $\mathcal{M}$.}
Then $\mathcal{M}$  gives the probability of each word $v$ in the vocabulary being filled {in the ``\texttt{[MASK]}" location}.
To map the probabilities of label words to the probabilities of labels, we define a verbalizer as a mapping $f$ from a few words in the vocabulary, which form the label-word set $\mathcal{V}$, to the label space $\mathcal{Y}$, i.e., $f: \mathcal{V} \mapsto \mathcal{Y}$.
The label-word set of class $y$ is denoted as $\mathcal{V}_y = (v_1,v_2,\dots,v_n)$, where  $\mathcal{V}_y$ is a subset of $\mathcal{V}$, i.e.,  $\cup\mathcal{V}_{y\in\mathcal{Y}}=\mathcal{V}$.

The probability of  predicting class  $y$ is computed as
\begin{equation}
	P(y\in\mathcal{Y}|\mathbf{x}_p) = g(P_\mathcal{M}( \texttt{[MASK]}=v|\mathbf{x}_{\text{p}}) |v\in \mathcal{V}_y ),
\end{equation}
where $g$ is a function employed to transform the probability of label words in $\mathcal{V}_y$  to the probability of the label $y$. If $P(y\in\mathcal{Y}|\mathbf{x}_p)$ is greater than a set threshold, the label $y$ is assigned to the input text $\mathbf{x}$.

\subsection{Verbalizer Construction}
The proposed verbalizer construction approach has two procedures: concept mining and cascade calibration.

\subsubsection{Concept Mining}
Concept mining is the procedure of mining rich scenario-specific concepts as label-word candidates. {The process comprises of two steps: named entity extraction, known as "named entity" for topic classification tasks and as "part-of-speech" for sentiment analysis tasks, and concept query. To streamline the process, in the subsequent sections, the extraction of both named entities and part-of-speech tags is unified and referred to as named entity extraction.}

\textbf{Named Entity Extraction} is the prerequisite work for concept query, which extracts multiple named entities \footnote{{In the context of topic classification, the extracted entities are referred to as named entities, while in the realm of sentiment analysis, they are identified as Part-of-Speech tags.}} from a task-specific scenario as query keys for a concept knowledge base. Multiple named entities serve as multiple perspectives to expand the coverage of a label-word space. Specifically, randomly selecting a set of training instances from a task-specific dataset as the support set $\mathcal{T}_n$. $n$  is the size of the set $\mathcal{T}_n$ and is not fixed, which depends on the {topic or sentiment polarity distribution} of different datasets.  Then all named entities extracted from $\mathcal{T}_n$ form {a named entity set  $\mathcal{N}_s$ } as the query keys for concept query.

\begin{table}
	\centering
	\resizebox{0.6\columnwidth}{!}{
		\begin{tabular}{lclc}
			\toprule
			{NE Type} &    & {NE Type} & \\
			\midrule
			PERSON & & LOCATION& \\
			ORGANIZATION & & MISC& \\
			CITY &  & COUNTRY& \\
			NATIONALITY& &RELIGION& \\
			TITLE& &CRIMINAL\_CHARGE&\\
			STATE\_OR\_PROVINCE & & CAUSE\_OF\_DEATH & \\
			\bottomrule
		\end{tabular}
	}
	\caption{The detail of the twelve selected NE tags used in {topic classification}.}
	\label{ner_tags}
\end{table}

\begin{table}
	\centering
	\resizebox{0.5\columnwidth}{!}{
		\begin{tabular}{lclc}
			\toprule
			POS Tag Type &    &POS Tag Type & \\
			\midrule
			ADV & & ADJ& \\
		
			\bottomrule
		\end{tabular}
	}
	\caption{{The detail of the two selected POS tags used in sentiment analysis.}}
	\label{pos_tags}
\end{table}

\textbf{Concept Query} obtains related concepts based on the named entity set $\mathcal{N}_s$  from an external concept base \citep{wuProbaseProbabilisticTaxonomy2012} as label-word candidate set $\mathcal{C}_v$. The whole procedure can be divided into three steps:
\begin{enumerate}
	\item Filter out irrelevant named entities from the candidate set $\mathcal{N}_s$. There are more than twelve types of named entities in $\mathcal{N}_s$. In this step, a fresh set $\mathcal{N}_{s'}$ is obtained by filtering out named entities that do not belong to the twelve selected types. 
	\begin{equation}
		\mathcal{N}_{s'} \leftarrow  \mathcal{N}_{s}
	\end{equation}
	{Please refer to Table \ref{ner_tags} for detailed information on the twelve selected Named Entity Recognition types utilized in topic classification tasks.}
	{For sentiment analysis tasks, a query key corresponds to a Part-of-Speech (POS) tag. For detailed information on the two selected POS types used in sentiment analysis tasks, please refer to Table \ref{pos_tags}.}
	\item Query related concepts based on the set  $\mathcal{N}_{s'}$. In the set, each item is a query key for the concept base \footnote{Probase from Microsoft Research}. {It denotes as}
	\begin{equation}
		\mathcal{C}_{v} \leftarrow  \mathcal{N}_{s'},
	\end{equation}
{ where $\mathcal{C}_{v}$ represents the set of candidate concepts that are obtained from the concept base.}
	\item Select concepts with a higher correlation-degree. Each quried concept has an associated probability value indicating the correlation-degree to its query key. To improve the cascade calibration procedure, we can refine the set $\mathcal{C}_{v}$ by selecting the top few items, the specific number can be determined by a pilot experiment \footnote{In our case, the number is 50.}.
	
\end{enumerate}

\subsubsection{Cascade Calibration}

Label-word candidates directly queried from an external concept base have noisy and irrelevant information and are over-covered in a label-word space.
For example, the concepts for the named entity ``APPLE" have two categories of ``fruit" and ``company", but for the label ``BUSINESS", ``fruit" is not a valid label word. In particular, for zero-shot based learning, the bias of a language model toward predicting certain answers \citep{zhaoCalibrateUseImproving2021c,kongCalibratedLanguageModel2020} is unavoidable and cannot be ignored. Therefore, the elimination of irrelevant or noisy candidates and the calibration of label words based on task-specific settings are essential.

To achieve the aforementioned refinement, we devise a cascade calibration procedure involving three steps: anchor creation, language model calibration, and category calibration.

\textbf{Anchor Creation}\quad
To calibrate a concept, a prerequisite is to prepare a base value, namely an anchor, to be employed to measure the fitness of a concept to a task-specific scenario.
The anchor is created based on a task-specific prompt template and a sample obtained from the task-specific dataset.

To create the anchor, we firstly compute the probablity of a token $t$  in the vocabulary $T$ of $\mathcal{M}$ at a masked location in a task-specific scenario as 
\begin{equation}
	P_t(\mathbf{x}_p,t)=P_\mathcal{M}(\texttt{[MASK]}=t|\mathbf{x}_\text{p}),
\end{equation}
where the prompt $\mathbf{x}_p$ denotes a sample wrapped into a task-specific prompt template.

Then for the convenience of computing the scenario-specific probability of each token $t$, we exploit normalized exponential function on each token $t$ by 

\begin{equation} 
	Q(t|\mathbf{x}_p)=\frac{\exp{P_t(\mathbf{x}_p,t)}}{\sum_{t'\in T}\exp{P_t(\mathbf{x}_p,t')}},
\end{equation}
where $t$ denots a specific item in $T$ to be computed, $t'$ stands for each word in $T$.

\textbf{Lanuguage Model Calibration}\quad
Among the candidate concepts for a verbalizer, there are still some concepts that are less likely than others to be predicted as eligible label words for a class by a PLM in a task-specific scenario. Concepts with low probability in task-specific scenarios need to be eliminated first.
Language model-based calibration is leveraged to remove anchor-based low-probability concepts.

Each candidate concept may not be exactly an item in the vocabulary of the PLM, therefore we define the function $S_c$  to compute the average probability of all tokens belonging to the concept as its predicted probability.

Equation \ref{eq_sc} is leveraged to compute the probability of the concept based on an anchor by
\begin{equation}
	\label{eq_sc}
	S_c(c,\mathbf{x},p)=  \frac{1}{|c|}\sum_{i=1 }^{|c|} 	Q(t_i|\mathbf{x}_p), 
\end{equation}
where $|c|$ stands for the number of tokens belonging to the concept $c$ and presenting in the vocabulary of the PLM, $p$ is the prompt template exploited in the task-specific scenario.
We score  each concept of $\mathcal{C}_v$ on all examples in support set $\mathcal{T}_v$ with
\begin{equation} 
	S_t(c,\mathcal{T}_v,p) = \sum_{x\in\mathcal{T}_v}S_c(c,x,p).
\end{equation}	
{The number of examples in the support set $\mathcal{T}_v$ utilized during the cascade calibration is a hyperparameter, denoted as $q$.}


The higher the score $S_t$, the higher the probability that a concept is selected as a label word. 
In the language model calibration step, we do not leverage the label information of samples.
In order to reduce the time consumption of computing the final scores for all concepts in the category calibration step, we select only top-$j$ concepts as the candidate concept set $\mathcal{C}_{v'}$.

\textbf{Category Calibration}\quad
For a label word that tends to predict a particular class label over others in a task-specific scenario, we need to select this word as the label word for the class. Thus, a calibration operation on each candidate based on the class label is essential and necessary.
Being inspired by \citep{dunningAccurateMethodsStatistics1993,schickExploitingClozeQuestionsFewShot2021}, we choose log-likelihood ratio as the score function, then select top-$l$ concepts as the label words for each class, in which $l$  is a hyperparameter and can be determined automatically and rapidly in a task pipeline as it has merely a narrow integer range.
The below log-likelihood ratio function can  measure the difference between likelihood $P_1$ and $P_2$, 
\begin{equation} 
	S_{llr}(P_1,P_2)=log\frac{P_1}{P_2},
\end{equation}	

where 
$P_1$ denotes the normalized exponent of a specific token $t$ and $P_2$ indicates the sum of counterpart for the rest tokens in the same vocabulary, therefore $P_1 + P_2 = 1$, and $P_1$ is expected greater than $P_2$.

Samples in the set $\mathcal{T}_v$ with the label $y$ are marked as positive samples stored in the set $\mathcal{T}_y$, ones with other labels as negative samples contained in the set $\mathcal{T}_{y'}$.
Before computing the score of a specific concept to be a qualified label word for the class $y$ on the task-specific scenario with $\mathcal{T}_y$, it is necessary to compute the score of each token belonging to the concept.
Our goal for a specific category is a series of one-vs-others selection tasks. Thus, the support set $\mathcal{T}_v$ is divided into two parts: $\mathcal{T}_y$ and $\mathcal{T}_{y'}$, i.e., $\mathcal{T}_{y'} = \mathcal{T}_v \setminus \mathcal{T}_y$ .

Considering merely one single token belonging to the concept which is selected as a label word, the probability of the token predicting the positive samples as class $y$ must be clearly greater than as other classes, we define the positive evaluating score $S_p$ as 

\begin{equation}\label{eq_posi_score}
	S_p(t,\mathcal{T}_y,p) = \sum_{\mathbf{x}\in \mathcal{T}_y}  log\frac{Q(t|\mathbf{x}_p)}{(1-Q(t|\mathbf{x}_p))}.
\end{equation}

Meanwhile, the probability that the token predicting the negative samples as other classes should be higer than as the class $y$, the negative evaluating score is defined as 

\begin{equation}
	S_n(t,\mathcal{T}_{y'},p) = \sum_{\mathbf{x}\in \mathcal{T}_{y'}}  log\frac{(1-Q(t|\mathbf{x}_p))}{Q(t|\mathbf{x}_p)}.
\end{equation}

For the token $t$,  its final score predicting class $y$ correctly is 
\begin{equation}
	S (t,\mathcal{T}_v,p,y)= S_p(t,\mathcal{T}_y,p) + S_n(t,\mathcal{T}_{y'},p).
\end{equation}

The concept $c$ is consists of several tokens, we take the averaged $S$ score of all tokens as the final score of the concept. To do so, we compute its score of predicting correctly samples  as class $y$  by 
\begin{equation}\label{eq12}
	S_{cv}(c,\mathcal{T}_v,y,p) = \sum_{i=1}^{|c|}S(t_i,\mathcal{T}_v,p,y),
\end{equation}
where $c = \{t_1,t_2,\dotsm,t_i\}$, $t_i$ denotes a token in the vocabulary of a PLM.

Equation \ref{eq12} is employed on each concept of the candidate set $\mathcal{C}_{v'}$ based on the support set $\mathcal{T}_v$ and a class label, then select top-$l$ concepts as the label-word set for the class. On a dataset of $k$ classes, for each prompt template $p$, we finally obtain a verbalizer of $k * l$ concepts.

\subsection{Verbalizer Utilization}
Verbalizers constructed using the ISCV approach are plug-and-play and can be directly applied in a task pipeline for prompt-tuning.
A verbalizer implements a mapping mechanism between a label-word space and a class-label space, that is, it projects the predicted probability of each final label-word set to the class label decision. A label word consists of several tokens in the vocabulary of a PLM. Conventionally, there are three ways to compute the probability of a label word in prompt-tuning: the value of the first token, the mean value of all token, or the max value of all tokens. We choose the mean value by pilot experiments. The mean probability of all concepts in a label-word set is chosen as the final value by investigating that the contribution of each label word of a class is equal. The predicted label $\hat{y}$ is
\begin{equation}
	\hat{y}=\operatorname{argmax}_{y\in\mathcal{Y}}\frac{\sum_{c\in\mathcal{V}_y}  \frac{ \sum_{t\in c}P_\mathcal{M}(\texttt{[MASK]}=t|\mathbf{x}_p)} {|c|} }{|\mathcal{V}_y|},
\end{equation}
where $\mathcal{V}_y$ denotes the label-word set of class $y$.
\section{Experiments}

We evaluate ISCV on three datasets with zero-shot text classification tasks to confirm the effectiveness of incorporating concepts into verbalizer construction.

\subsection{Experimental Settings}

\subsubsection{Datasets, Templates and Evaluation Metrics}
{Experiments are conducted on three topic classification datasets: AG's News \citep{zhangCharacterlevelConvolutionalNetworks2015b}, DBPedia \citep{lehmannDBpediaLargescaleMultilingual2015} and Yahoo \citep{zhangCharacterlevelConvolutionalNetworks2015b}. To confirm its effectiveness on other tasks, experiments are also conducted on two sentiment analysis datasets: Amazon \citep{mcauleyHiddenFactorsHidden2013} and IMDB \citep{maasLearningWordVectors2011}. }
We examine the zero-shot based prompt-tuning method with four hand-crafted templates and report both average results with standard deviation and the best results of the templates, the best results are in brackets. For a fair comparison, all experiments exploit same manual templates. Manual templates are confirmed to be comparable to auto-generated templates \citep{gaoMakingPretrainedLanguage2021a} as they contain abundant expert knowledge  and have excellent explainability.

Following previous works, we adopt Micro-F1 as the metric for all experiments. Each result is obtained experimentally with the same random seed as in the previous work \citep{huKnowledgeablePrompttuningIncorporating2022}.

\begin{table}
	\centering
	\resizebox{0.6\columnwidth}{!}{
		\begin{tabular}{llll}
			\toprule
			{Dataset Name}	& \ Class & Train Size & Test Size \\
			\midrule
			AG's News & 4 &  120000 &7600 \\
			
			DBPedia & 14 &  560000 &70000 \\
			Yahoo  & 10 & 1400000& 60000 \\
			{Amazon}  & {2} & {3600000}& {10000} \\
			{IMDB} &{2} & {25000}& {25000} \\
			\bottomrule
		\end{tabular}
	}
	\caption{{The statistics of all datasets. The original dataset, Amazon, has a test set consisting of 400,000 samples. For the utilized test set of dataset Amazon, ISCV randomly samples 10,000 samples. For fair comprison, KPT is also reproduced on the same test set. }}
	\label{datasetdetail}
\end{table}

\subsection{Details of Datasets and Templates}
\label{sec:details_templates}

 {ISCV is a novel approach that aims at the construction of verbalizers for text classification tasks: topic classifcation and sentiment analysis.} {We conduct topic classification experiments on three topic classification datasets: AG's News \citep{zhangCharacterlevelConvolutionalNetworks2015b}, DBPedia \citep{lehmannDBpediaLargescaleMultilingual2015} and Yahoo \citep{zhangCharacterlevelConvolutionalNetworks2015b}. Sentiment analysis experiments are also conducted on Amazon \citep{mcauleyHiddenFactorsHidden2013} and IMDB \citep{maasLearningWordVectors2011} datasets to confirm the effectiveness of ISCV.} {The statistics of all datasets are presented in Table \ref{datasetdetail}}.
{Our goal is to construct an effective verbalizer of a dataset for prompt-tuning based classification tasks.} For the objective, designing or constructing a prompt template is orthogonal to our approach. For a fair comparison with KPT \citep{huKnowledgeablePrompttuningIncorporating2022} or alternative baselines, we choose templates offered in OpenPrompt, which are the same as in KPT, each dataset has four manual templates. 
The prompt templates for each dataset are described in the following sections.

\textbf{AG's News} is a dataset used for news topic classification. The text piece of each sample is denoted by $\mathbf{x}$. The four templates used in this dataset are listed in the following text frame:

\begin{center}
	\begin{tcolorbox}[colback=gray!10,
		colframe=black,
		width=8cm,
		arc=1mm, auto outer arc,
		boxrule=0.5pt,
		]
		\begin{align*}
			T_1(\mathbf{x})  &= \text{ A \texttt{[MASK]} news : }  \mathbf{x}  \\
			T_2(\mathbf{x})  &= \mathbf{x} \text{ This topic is about \texttt{[MASK]}. }\\
			T_3(\mathbf{x})   &= \text{ [ Category : \texttt{[MASK]} ] }  \mathbf{x}  \\
			T_4(\mathbf{x})   &=\text{ [ Topic : \texttt{[MASK]} ] } \mathbf{x} 
		\end{align*}
	\end{tcolorbox}
\end{center}

\textbf{DBPedia} is a dataset used for topic classification. Each sample in DBPedia contains two pieces of text: the title and the content. Within the following templates, $\mathbf{a}$ and $\mathbf{b}$ refer to the title and content of a sample, respectively. Furthermore, $\tilde{\mathbf{a}}$ denotes the title without its final punctuation mark.
\begin{center}
	\begin{tcolorbox}[colback=gray!10,
		colframe=black,
		width=8.5cm,
		arc=1mm, auto outer arc,
		boxrule=0.5pt,
		]
		\begin{align*}
			T_1(\mathbf{a},\mathbf{b}) &=  \mathbf{a} \ \mathbf{b}  \  \tilde{\mathbf{a}}  \text{ is a \texttt{[MASK]} . }\\
			T_2(\mathbf{a},\mathbf{b}) &=  \mathbf{a} \   \mathbf{b}  \  \text{ In this sentence, } \  \tilde{\mathbf{a}}  \text{ is a \texttt{[MASK]} . } \\
			T_3(\mathbf{a},\mathbf{b}) &=   \mathbf{a}\ \mathbf{b} \  \text{The type of}\  \tilde{\mathbf{a}} \ \text{is}\  \texttt{[MASK]} .\\
			T_4(\mathbf{a},\mathbf{b}) &=\mathbf{a}\ \mathbf{b} \ \text{The category of}\  \tilde{\mathbf{a}} \ \text{is}\ \texttt{[MASK]} . 
		\end{align*}
	\end{tcolorbox}
\end{center}

\textbf{Yahoo} is a question dataset obtained from yahoo.com, used for topic classification. The complete name of this dataset is \textbf{yahoo\_answers\_topics}. The prompt templates used for this dataset are identical to those used for the AG's News dataset, with the only difference being the replacement of the word "news" with "question". {The text piece of each sample is denoted by $\mathbf{x}$. }

\begin{center}
	\begin{tcolorbox}[colback=gray!10,
		colframe=black,
		width=8cm,
		arc=1mm, auto outer arc,
		boxrule=0.5pt,
		]
		\begin{align*}
			T_1(\mathbf{x})  &= \text{ A \texttt{[MASK]} question : }  \mathbf{x}  \\
			T_2(\mathbf{x})  &= \mathbf{x} \text{ This topic is about \texttt{[MASK]} . }\\
			T_3(\mathbf{x})   &= \text{ Category : \texttt{[MASK]} }  \mathbf{x}  \\
			T_4(\mathbf{x})   &=\text{ Topic : \texttt{[MASK]} } \mathbf{x}  
		\end{align*}
	\end{tcolorbox}
\end{center}


{The \textbf{Amazon} dataset is utilized for sentiment analysis and comprises several million customer reviews from Amazon. Based on this dataset, we define the following templates:}
\begin{center}
	\begin{tcolorbox}[colback=gray!10,
		colframe=black,
		width=8cm,
		arc=1mm, auto outer arc,
		boxrule=0.5pt,
		]
		\begin{align*}
			T_1(\mathbf{x})  &= \text{ It was \texttt{[MASK]} . }  \mathbf{x}  \\
			T_2(\mathbf{x})  &=  \text{ Just \texttt{[MASK]} ! } \mathbf{x}\\
			T_3(\mathbf{x})   &=  \mathbf{x} \text{ All in all, it was \texttt{[MASK]} . }   \\
			T_4(\mathbf{x})   &= \mathbf{x} \text{ In summary, it was \texttt{[MASK]} . }  
		\end{align*}
	\end{tcolorbox}
\end{center}

{, where $\mathbf{x}$ represents the textual content of a sample.}

{The \textbf{IMDB} dataset is another commonly used dataset for sentiment analysis, specifically for analyzing movie reviews. Following the templates defined in Schick et al. (2021) for sentiment analysis, we have established the following templates:}
\begin{center}
	\begin{tcolorbox}[colback=gray!10,
		colframe=black,
		width=8cm,
		arc=1mm, auto outer arc,
		boxrule=0.5pt,
		]
		\begin{align*}
			T_1(\mathbf{x})  &= \text{ It was \texttt{[MASK]} . }  \mathbf{x}  \\
			T_2(\mathbf{x})  &=  \text{ Just \texttt{[MASK]} ! } \mathbf{x}\\
			T_3(\mathbf{x})   &=  \mathbf{x} \text{ All in all, it was \texttt{[MASK]} . }   \\
			T_4(\mathbf{x})   &= \mathbf{x} \text{ In summary, the film was \texttt{[MASK]} . }  
		\end{align*}
	\end{tcolorbox}
\end{center}
{, where $\mathbf{x}$ represents the textual content of a sample.}

\subsubsection{Implementation Details}

All experiments in this study are conducted using OpenPrompt \citep{ding2021openprompt}, an open-source framework designed for prompt-based learning. As per the precedent set by prior works, we employed $\text{RoBERTa}_\text{large}$ \citep{liuRoBERTaRobustlyOptimized2019a} as the pre-trained language model for all experiments.

The ISCV framework involves three primary hyperparameters. The first hyperparameter, represented by $n$, corresponds to the number of samples retrieved as named-entity sources during the Named Entity Extraction procedure. After conducting a pilot experiment, we set this hyperparameter to 4,000.
The second hyperparameter, represented by $j$, denotes the number of top-$j$ concepts selected during the Language Model Calibration stage. We set this hyperparameter to 10,000, as it achieves a balance between effectiveness and computational consumption. This value was determined through pilot experiments conducted on {all datasets}.
Finally, during the Category Calibration procedure, the hyperparameter $l$ represents the top-$l$ concepts chosen as the final label words for each class.

Additional hyperparameters, such as truncated length and random seeds, are listed in Table \ref{tab:other-hyper-parameter}.

\begin{table}
	\centering
		\begin{tabular}{lcc}
			\toprule
			Parameter & Dataset  & Value \\
			\midrule
			Trunc Len & AG's News, DBPedia, Yahoo & 128 \\
		{Trunc Len} & {Amazon, IMDB} & {512} \\
			{Seeds for Zero-shot}& All & 144,145,146 \\
			{Seeds for Few-shot}& {AG's News, DBPedia, Yahoo} & {144,145,146,147,148} \\
			\bottomrule
		\end{tabular}
	\caption{The settings for additional hyperparameters. Trunc Len refers to the truncated length of an input text.}
	\label{tab:other-hyper-parameter}
\end{table}

\subsection{Baselines}
In this section, we introduce the three recent and stronge baselines in our evaluation. 

\textbf{Prompt-tuning (PT)} \quad Regular prompt-tuning methods typically rely solely on the class name as the label word for each class, as demonstrated by PET \citep{schickExploitingClozeQuestionsFewShot2021}. In our study on zero-shot prompt-based learning, we disregard any techniques from PET and other studies that are not pertinent to our contribution.

\textbf{Prompt-tuning with contextual calibration (PT+CC)} \quad  In \citep{huKnowledgeablePrompttuningIncorporating2022}, the concept of contextual calibration was proposed to significantly enhance the performance of regular prompt-tuning methods. Given its substantial improvement over regular prompt-tuning, the combination of regular prompt-tuning and contextual calibration is considered a new baseline for comparison purposes.


\textbf{Knowledgeable prompt-tuing (KPT)}\quad KPT \citep{huKnowledgeablePrompttuningIncorporating2022} is an excellent method that utilizes external knowledge to construct verbalizers. It avoids the need to expand synonyms of class names as label words, and instead enhances the label-word space perspective by incorporating related words of class names, thereby expanding the coverage of the label word space in a verbalizer. Through the use of context calibration, relevance refinement, frequency refinement, and learnable refinement to construct valid label words, KPT achieves superior performance compared to its prior works.

\subsection{Experimental Results}
\subsubsection{Main Experiments}

\label{main_experiment}
All experimental results on the {five} datasets are in Table \ref{main_result}.

\begin{table*}[!th]
	\centering
	\resizebox{1.0\textwidth}{!}{
		\begin{tabular}{lcccccc}
			\toprule
			Method&   AG's News & DBPedia & Yahoo & {Amazon} & {IMDB}\\
			\midrule
			PT \citep{huKnowledgeablePrompttuningIncorporating2022} & 75.1$\pm$6.2  {(79.0)} &  66.6$\pm$2.3 {(68.4)} & 45.4$\pm$7.0 {(52.0)} & \textcolor{red}{ 80.2$\pm$8.8{(87.8)} } &  86.4$\pm$4.0{(92.0)} \\
			PT+CC \citep{huKnowledgeablePrompttuningIncorporating2022}& 79.9$\pm$0.7 {(81.0)} & 73.9$\pm$4.9 {(82.6)} & 58.0$\pm$1.4 {(58.8)}  & \textcolor{red}{91.4$\pm$1.6{(93.5)}} & 91.6$\pm$3.0{(93.7)} \\
			
			KPT \citep{huKnowledgeablePrompttuningIncorporating2022}& 84.8$\pm$1.2 {(86.7)} & 82.2 $\pm$5.4 {(87.4)}&  61.6 $\pm$2.2 {(63.8)}  & \textcolor{red}{92.8$\pm$1.2{(94.6)}}  & 91.6$\pm$2.7{(94.0)}\\
			\midrule
			
			PT \citep{huKnowledgeablePrompttuningIncorporating2022}* &  75.1$\pm$\bigcolor{5.6}{(79.0)} &  66.6$\pm$\bigcolor{2.1} {(68.4)} & 45.4$\pm$\smallcolor{6.4}{(52.0)}  & \smallcolor{81.1}$\pm$\smallcolor{7.5}\smallcolor{(88.0)} & {86.4}$\pm$\smallcolor{3.6}{(92.0)}\\
			PT+CC \citep{huKnowledgeablePrompttuningIncorporating2022}* &79.9$\pm$0.7{(81.0)} & \bigcolor{75.1}$\pm$\bigcolor{5.3} (\smallcolor{82.7}) &\bigcolor{57.9}$\pm$\smallcolor{1.5}(\smallcolor{59.0})& \smallcolor{91.2}$\pm$\smallcolor{1.5}\smallcolor{(93.6)}&  \smallcolor{90.6}$\pm$\smallcolor{3.2}\smallcolor{(93.8)} \\
			
			KPT  \citep{huKnowledgeablePrompttuningIncorporating2022}*&  84.8$\pm$1.2(\smallcolor{86.9}) & 82.2$\pm$5.4 (\smallcolor{87.5})&  \bigcolor{61.8}$\pm$\bigcolor{2.0}(\smallcolor{64.0})  &   \bigcolor{92.4}$\pm$\smallcolor{1.3}(\smallcolor{94.3}) & {91.6}$\pm${2.7}(\smallcolor{94.1}) \\
			\midrule
			
			ISCV  &   \textbf{ 87.3}$\pm$\textbf{0.4} \small\textbf{(87.8)}& \textbf{90.4}$\pm$\textbf{1.2} (\textbf{91.6})& \textbf{65.3}$\pm$\textbf{0.8} (\textbf{66.6})  &\textbf{92.7}$\pm$\textbf{1.6} (\textbf{94.5}) & \textbf{91.9}$\pm$\textbf{3.3} (\textbf{94.7}) \\

			\bottomrule
		\end{tabular}
	}
	\caption{{Results on zero-shot based topic classification and sentiment analysis tasks.} The best results for the templates are shown in parentheses. The baseline results marked with * are obtained experimentally, and the other results are quoted from the paper of KPT. All results of ISCV method achieve new state-of-the-art in \textbf{bold}. The data with \underline{underline} indicates it altered relative to the quoted result. Each result is the average of a total of 12 experiments on four prompt templates and three random seeds for a dataset. {KPT initially selected a test set consisting of 10,000 samples from the test set of the Amazon dataset. However, unfortunately, the test set was subsequently lost by KPT. The red result denotes that it was obtained by utilizing the undisclosed test set cited from the research paper of KPT. ISCV performs random sampling of 10,000 examples from the test set of the Amazon dataset and reproduces the findings of KPT.}}
	\label{main_result}
\end{table*}

{Compared to all baseline methods, ISCV achieves fresh state-of-the-art results on all datasets, especially on the three topic classification datasets.} On the datasets AG News, DBPedia, and Yahoo, the error rate was reduced by an average of 2.5\%, 8.2\%, and 3.5\%, respectively. The standard deviation indicates the stability of a method on a dataset.  In comparison with the results obtained experimentally with KPT, the stability is improved by 0.8\%, 4.2\%, and 1.2\% on all three datasets, respectively. The PT+CC method has the best stability among the three baselines, and ISCV still outperforms it by 0.3\%, 0.8\%, and 0.4\% on stability alone. The maximum Micro-F1 value for each template on all three datasets still emerges from the ISCV results.
{On the sentiment analysis datasets of Amazon and IMDB, the ISCV method achieves comparable yet superior results. Given the substantial advantages observed in the results obtained from the three topic classification datasets, our next step involves analyzing the underlying reasons behind these outcomes.}  


Due to leveraging a large number of concepts with higher-level abstraction as label-word candidates, ISCV method improves the coverage of the label-word space in a verbalizer. With the help of concept calibration based on scenario-specific information, ISCV reduces the bias of the label-word space . 

Our ISCV approach outperforms all baselines in all experiments on the task of zero-shot text classification in prompt-based learning, confirming the effectiveness of incorporating scenario-specific concepts into the verbalizer construction in prompt-tuning. {Note that our approach achieves new stable and state-of-the-art performance on all three topic datasets.}

\subsubsection{Stability of Templates}

\begin{figure*}[h]
	\centering
	\includegraphics[width=1\textwidth]{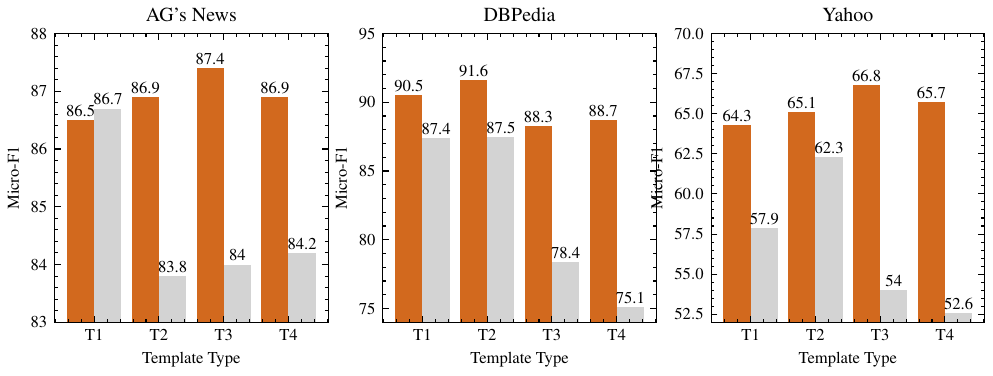} 
	\caption{
		Comparison between ISCV and KPT for template stability. The orange color indicates ISCV and the gray color indicates KPT.  T1, T2, T3, and T4 denote the four templates for each dataset separately and each number at the top of the colored bar is the value of the corresponding Micro-F1 value.
	}
	\label{stability}
\end{figure*}

The experimental results of template stability are shown in Figure \ref{stability}.
Details of the templates for each dataset are given in Section \ref{sec:details_templates}. For fair comparison, all results are obtained experimentally based on the same hyperparameters; See Table \ref{tab:other-hyper-parameter} for other relevant hyperparameters.

By investigation, template stability depends on the distribution of label words on all classes of a dataset. Since KPT employs a same set of fixed number of words as candidates for all templates on a dataset, the candidate space for label words is not sufficient as different templates elicit different knowledge from the same PLM, and hence a different set of label-word candidates is required. 
The probability distribution of each word in the vocabulary for a PLM  responding different prompt templates are different \citep{reynoldsPromptProgrammingLarge2021a} correlated  directly to a specific template.

ISCV confirms that a larger label-word candidate space, different prompt templates, and scenario-specific label-word calibration can be used to expand the coverage of a label-word space and improve stability across different templates to construct more robust verbalizers.

\begin{figure*}[!h]
	\centering
	\includegraphics[width=1\textwidth]{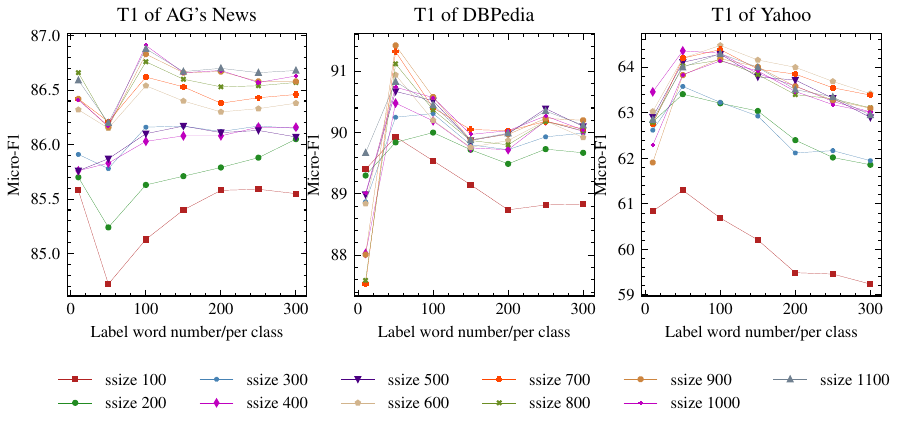} 
	\caption{
		Parameter Sensitivity of ISCV on each template. ``ssize" denotes the size of a support set.
	}
	\label{fig:sensitive}
\end{figure*}

\subsubsection{Parameter Sensitivity}


The present study focuses on two crucial hyperparameters, namely the size of the support set $\mathcal{T}_v$ and the number of label words for each class on a dataset. To streamline the investigation, both hyperparameters are evaluated using the first prompt template of each dataset.

The experimental results are shown in Figure \ref{fig:sensitive}. 

\begin{table}
	\centering
	\begin{tabular}{lcc}
		\toprule
		Dataset  & Support Set & Label Words\\
		\midrule
		AG's News  & 1100  & 100\\
		DBPedia & 800  & 50\\ 
		Yahoo& 600  & 100\\
		{Amazon}& 	{900}  & 	{250}\\
		{IMDB}& 	{1200}  & 	{50}\\
		
		\bottomrule
	\end{tabular}
	\caption{
		{Sample number of support set ($\mathcal{T}_v$) and number of label words for each class for all datasets. The number of label words indicates the number of concepts selected as label words for each class of a dataset.}
	}
	\label{tab:hyperparameter_set}
\end{table}

The size of the support set $\mathcal{T}_v$, which refers to the number of samples in $\mathcal{T}_v$, has a significant impact on the coverage of a label-word space. Based on Figure \ref{fig:sensitive}, it is evident that various datasets require different support set sizes. For instance, the construction of a verbalizer for dataset AG's News requires 1000 support examples to achieve promising results, while dataset DBPedia and Yahoo require 800 and 600 support instances, respectively. The final size of the support set is determined by the average Micro-F1 value across all templates on a given dataset, as presented in Table \ref{tab:hyperparameter_set}. This phenomenon can be attributed to the varying diversity of topics among different datasets, which necessitates a distinct number of samples to achieve optimal results.
Furthermore, a positive correlation exists between the results and the size of the support set $\mathcal{T}_v$.

Number of labeled words on a dataset for each class. It requires only a small number of label words per class to achieve promising results. Over all three datasets, 100 or fewer concepts per class is sufficient to achieve decent results. The reason behind the results is that each concept has a higher level of abstraction, yielding a wider coverage than the label words obtained in previous studies.

The above experiments demonstrate that both hyperparameters have a profound effect on the label-word space of a verbalizer, but there is a specific threshold for each hyperparameter in a task-specific scenario.

\subsection{Ablation Experiments}


\begin{table*}[!th]
	\centering
	\resizebox{1.0\textwidth}{!}{
		\begin{tabular}{lcccc}
			\toprule
			Method&  	& AG's News & DBPedia & Yahoo \\
			\midrule
			
			
			ISCV  &  & \textbf{87.3} $\pm$\textbf{0.4}\small\textbf{(87.8)}& \textbf{90.4} $\pm$\textbf{1.2}(\textbf{91.6})& \textbf{65.3 }$\pm$\textbf{0.8}(\textbf{66.6})   \\
			\midrule
			
			KPT \citep{huKnowledgeablePrompttuningIncorporating2022}&  &84.8$\pm$1.2{(86.7)} & 82.2$\pm$5.4{(87.4)}&  61.6 $\pm$2.2{(63.8)}  \\
			\midrule
			
			$\text{ISCV}_{LM+KPT}$&  &86.5$\pm$0.7(87.2) & 86.2$\pm$ 4.8(91.8) & 64.7$\pm$ 1.8(66.5)  \\
			\midrule
			$\text{ISCV}_{KPT}$&  &66.8$\pm$0.6(67.6) & 43.2$\pm$ 6.0(51.1) & 38.0$\pm$ 4.6(42.2)   \\
			\bottomrule
		\end{tabular}
	}
	\caption{Ablation results on zero-shot based classification tasks. The best results for the templates are shown in parentheses.}
	\label{tab:ablation_result}
\end{table*}

To demonstrate the effectiveness of the cascade calibration in our proposed ISCV, we design a set of ablation experiments on zero-shot based text classification. The experimental results are presented in Table \ref{tab:ablation_result}. The proposed ISCV method consists of two procedures: concept mining and cascade calibration. Cascade calibration consists of three steps: anchor creation, language model calibration, and category calibration. Concept mining is the source of label word candidates and it cannot be replaced with other procedures. {In Table \ref{tab:ablation_result}}, the first row shows the results for the full ISCV and the second row for the full KPT \citep{huKnowledgeablePrompttuningIncorporating2022}.
{All remaining experiments utilize the result of concept mining as the source of label word candidates. The third row,  $\text{ISCV}_{LM+KPT}$, corresponds to the experiment where the category calibration of ISCV is replaced by the calibration method of KPT. The fourth row, 
$\text{ISCV}_{KPT}$, corresponds to the experiment where the whole cascade calibration is replaced by the counterpart of KPT.}

From Table \ref{tab:ablation_result}, we can observe that replacing the cascade calibration of ISCV with the calibration method of KPT yields only the worst results. Replacing the category calibration step of ISCV with the calibration method of KPT can still outperform the baseline KPT. { The ablation results demonstrate that the entire cascade calibration is an integral procedure. Anchor creation provides a foundation for subsequent calibration steps. Language model calibration helps eliminate noise that is irrelevant to the scenario-specific setting within the candidate words, which cannot be effectively denoised by other methods, such as the counterpart of KPT. Furthermore, category calibration also yields better improvements compared to the calibration part of KPT. }

The results of all ablation experiments presented in Table \ref{tab:ablation_result} confirm the effectiveness of our proposed ISCV approach, with both language model calibration and category calibration demonstrating significant effectiveness.

\subsection{Few-shot Learning}

\begin{table*}
	\centering
	\resizebox{0.7\textwidth}{!}{
		\begin{tabular}{lllll}
			\toprule 
			Shot  & Method&  AG's News & DBPedia &Yahoo \\
			\midrule
			\multirow{2}{*}{1}  &  \textsc{KPT} &  83.7$\pm$3.5(84.6) & 93.7$\pm$1.8(95.3)& 63.2$\pm$2.5(64.1) \\
			&  \textsc{ISCV}   					&  \textbf{83.8$\pm$1.9(86.5)} & \textbf{95.3$\pm$1.3(97.0)} & \textbf{65.2$\pm$2.0(67.7)}\\
			\midrule
			\multirow{2}{*}{5} &\textsc{KPT}     &  85.1$\pm$1.0(85.8)  &97.2$\pm$0.4(97.3)  & \textbf{67.2$\pm$0.8(67.8)}  \\			& \textsc{ISCV}						 &  \textbf{85.7$\pm$2.6(88.3)}  &  \textbf{97.3$\pm$0.5(98.1)}   & 67.1$\pm$0.6(68.2)\\
			\midrule
			
			\multirow{2}{*}{10} & \textsc{KPT}     &  86.3$\pm$1.6(87.0)  & \textbf{98.0$\pm$0.2(98.1)}   & \textbf{68.0$\pm$0.6(68.2)} \\
			
			& \textsc{ISCV}							&  \textbf{87.0$\pm$1.4(88.9)}  &  97.9$\pm$0.4(98.4) & 67.9$\pm$0.3(68.6)\\
			
			\midrule
			\multirow{2}{*}{20} & \textsc{KPT}     &87.7$\pm$0.6(87.8)  &98.1$\pm$0.3(98.2)  & \textbf{69.0$\pm$0.7(69.3) }\\
				
			& \textsc{ISCV}							& \textbf{88.4$\pm$0.6(89.3)}  & \textbf{98.1$\pm$0.2(98.5)}  & 68.8$\pm$0.6(69.5)\\
			
			\bottomrule
		\end{tabular}
	}
	\caption{
		Results for few-shot setting. ISCV denotes our method. KPT refers to the best results of the baseline. A bold result indicates the best result with the same dataset and number of shots. The best results for the templates are shown in parentheses. 
	}
	\label{tab:fewshot-iscv-kpt}
\end{table*}

This paper aims to incorporate scenario-specific concepts into a verbalizer for zero-shot text classification. However, to validate its effectiveness in a few-shot setting, we design a set of few-shot experiments for prompt learning-based text classification.
To ensure a fair comparison, we select shot 1, shot 5, shot 10, and shot 20 as experimental settings \citep{huKnowledgeablePrompttuningIncorporating2022}. The results of the few-shot experiments are presented in Table \ref{tab:fewshot-iscv-kpt}, which demonstrate that the few-shot results of our proposed ISCV are superior to the previous best baselines in the shot-1 setting and comparable to the previous best baselines in other settings. Prior research \citep{huKnowledgeablePrompttuningIncorporating2022} suggests that calibration is not critical for the few-shot setting, as the need for calibration is significantly reduced with supervised input data, which is also supported by the results in Table \ref{tab:fewshot-iscv-kpt}.

\subsection{Analysis of Constructed Verbalizer}

{ISCV is proposed as an approach for constructing verbalizers that are specifically designed for zero-shot-based text classification tasks, such as topic classification and sentiment analysis. In the following sections, we provide a detailed analysis of the constructed verbalizer for each of these tasks.}

\subsubsection{{Topic Classifcation}}

Based on the findings presented in Table \ref{verbalizer-tab-topic}, it is evident that all the final label words correspond to relevant concepts. For instance, when examining the label ``POLITICS" from the AG's News dataset, we observe that there is overlap and expansion between the concepts of "national terrorist" and "international terrorist." This indicates reduced voids within the label-word space of the constructed verbalizer, resulting in a label-space that exhibits higher coverage and more continuity. It is worth noting that the cascade calibrated verbalizer demonstrates improved coverage and reduced bias.

\begin{table*}
	\centering
	\resizebox{\textwidth}{!}{
		\begin{tabular}{lll}
			\toprule 
			Dataset  & Label &  Label Words \\
			\midrule
			\multirow{2}{*}{AG's News}  &  \textsc{politics} & national terrorist, international terrorist, international conflict, international, international celebrity ...\\
			&  \textsc{sports}   & sports, league championship, league coach, league baseball team, football coach, league team ... \\
			\midrule
			\multirow{2}{*}{DBPedia}      &\textsc{corporation}     & software company, public company, public corporation, service company, consumer company ...  \\
			& \textsc{school}& campus facility, campus university, private nonprofit institution, private institute, college campus ... \\
			\midrule
			
			\multirow{2}{*}{Yahoo} & \textsc{society}     & atheistic religion, atheist religion, world religion, popular religion, religion most major religion ...\\
			
			& \textsc{science}&science, scientific study, natural scientist, scientific technique, scientific expert, field scientist ... \\
			
			\bottomrule
		\end{tabular}
	}
	\caption{Example label words for topic classification constructed with ISCV. Yahoo denotes dataset \textbf{yahoo\_answers\_topics}. For demonstration purposes, only two classes are presented per dataset.}
	\label{verbalizer-tab-topic}
\end{table*}

\subsubsection{{Sentiment Analysis}}

\begin{table*}
	\centering
	\resizebox{\textwidth}{!}{
		\begin{tabular}{lll}
			\toprule 
			Dataset  & Label &  Label Words \\
			\midrule
			\multirow{2}{*}{Amazon}  &  \textsc{negative} & paper garbage, gross error, badge, market failure, terrible mistake, wrong food, negative opinion, limited negative, ...\\
			&  \textsc{positive}   & love, great instrumental, good compliment, great champion, great benefit, order pleasure, member incredible value,... \\
			\midrule
			\multirow{2}{*}{IMDB}      &\textsc{negative}     & horrible problem, garbage, rap, real debacle, hollow, terrible mistake, owl, basic no no, horrible name, stupid joke, ...  \\
			& \textsc{positive}& rich feature, music legend, play, good compliment, time honored classic, member incredible value, form of love, ... \\

			\bottomrule
		\end{tabular}
	}
	\caption{Example label words for sentiment analysis constructed with ISCV.}
	\label{verbalizer-tab-analysis}
\end{table*}

Upon examining Table \ref{verbalizer-tab-analysis}, it becomes apparent that the final label words constructed by ISCV for sentiment analysis tasks encompass concepts with polarity. ISCV employs a Part-of-Speech task to extract relevant adjective and adverbial words, which serve as key queries to retrieve concepts from a concept base. Subsequently, through cascade calibration, ISCV selects concepts with positive polarity as label words for the positive class of a dataset, and vice versa for the negative class.

\subsection{{Comparison of ISCV with Non-Prompt Approaches}}

{The primary objective of the ISCV approach is to construct verbalizers specifically tailored for prompt-tuning in zero-shot text classification tasks. Previous studies, such as LOTClass \citep{mengTextClassificationUsing2020a}, have introduced methods that combine the output of the masked language model head with class names to enhance the label words for text categories, without relying on prompt tuning. This approach exhibits applicability in both zero-shot and few-shot text classification scenarios. To compare the effectiveness of ISCV with non-prompt tuning methods, we conducted experiments on four datasets: AG's News, DBPedia, IMDB, and Amazon. The specific details of these datasets are presented in Table \ref{datasetdetail}. Experiments conducted on datasets other than Amazon were evaluated using their respective original test sets.
The LOTClass method is set as the baseline, and zero-shot and few-shot text classification experiments are conducted utilizing the pre-trained language model bert-base-uncased \footnote{https://huggingface.co/bert-base-uncased}.}

\begin{table*}
	\centering
	\resizebox{\textwidth}{!}{
		\begin{tabular}{llcccc}
			\hline Model & Methods & AG's News & DBPedia & IMDB & Amazon \\
			\hline 		
			\multirow{2}{*}{ BERT } 
			& LOTClass w/o. self train \citep{mengTextClassificationUsing2020a} & 0.822 & 0.860 & 0.802 & $0.853(400K)$ \\
			& {ISCV}$_{BERT}$ & $\mathbf{0.865}$ & $\mathbf{0.881}$ & $\mathbf{0.805}$ & $\mathbf{0.863}(400K)$ \\
			
			\hline 
			{ RoBERTa  } & ISCV & ${0.873}$ & ${0.904}$ & ${0.919}$ & ${0.927}(400K)$ \\
			
			\hline
		\end{tabular}
	}
	
	\caption{{The results for no-self-trainig/zero-shot approaches. The ``BERT" model, which refers to the bert-base-uncased model, and the ``RoBERTa" model, which refers to the $\text{RoBERTa}_\text{large}$ model. The best result for each dataset based on the BERT model is indicated by highlighting it in \textbf{bold}. The size of the test set used for the Amazon dataset is denoted by the number in parentheses, where ``$K$" represents a thousand (1000). The notation $\text{ISCV}_{BERT}$ is used to represent the zero-shot-based  ISCV method employing the bert-base-uncased model.}}
	\label{comparison_result_of_non_prompt_zeroshot}
\end{table*}

\subsubsection{{Zero-Shot Experiments}}

{
The zero-shot experiments conducted on the Amazon dataset utilized its original test set, which comprised 400,000 examples.  Compared to the zero-shot experiments conducted with ISCV, the baseline method is denoted as ``LOTClass w/o (without) self-training". In the zero-shot or no-self-train experiments, we freeze the parameters of the bert-base-uncased model and did not perform any fine-tuning on it.
By referring to Table \ref{comparison_result_of_non_prompt_zeroshot}, it is evident that the ISCV method based on the bert-base-uncased model ($\text{ISCV}_{BERT}$) achieves superior results compared to the baseline. In particular, the $\text{ISCV}_{BERT}$ method has demonstrated remarkable improvements on topic classification datasets, with an increase of 4.3\% and 2.1\% observed on the AG's News and DBPedia datasets, respectively. 
On the sentiment analysis datasets, IMDB and Amazon, the $\text{ISCV}_{BERT}$ method exhibits a slight performance advantage over LOTClass, with improvements of 0.3\% and 1.0\%, respectively. The results undeniably demonstrate the advantages of prompt-tuning based methods, particularly in the context of topic classification tasks.}

{
When comparing ISCV and $\text{ISCV}_{BERT}$, the key difference lies in the underlying model used. ISCV relies on the $\text{RoBERTa}_\text{large}$ model, whereas $\text{ISCV}_{BERT}$ utilizes the bert-base-uncased model.
We observed that the ISCV method outperforms $\text{ISCV}_{BERT}$ by 0.8\%, 2.3\%, 11.4\%, and 6.4\% on datasets AG's News, DBPedia, IMDB, and Amazon, respectively.
These findings provide compelling evidence that prompt-tuning based zero-shot methods exhibit tremendous potential when applied to larger pre-trained language models.
Prompt-tuning benefits from the knowledge retrieved from a language model, and $\text{RoBERTa}_\text{large}$, with its larger number of parameters and extensive pre-training on diverse language materials\footnote{https://huggingface.co/roberta-large, https://huggingface.co/bert-base-uncased}, provides additional advantages in this regard \citep{liuPretrainPromptPredict2021a}.
}

\begin{table*}
	\centering
	\resizebox{\textwidth}{!}{
		\begin{tabular}{lllcccc}
			\hline Model & Shots & Methods & AG's News & DBPedia & IMDB & Amazon \\
			\hline 		
			\multirow{2}{*}{ BERT } 
			& Full  &LOTClass \citep{mengTextClassificationUsing2020a} & $\underline{0.864}$ & $\underline{0.911}$ & $\underline{0.865}$ & ${0.916}(400K)$ \\
			& Full  &LOTClass* \citep{mengTextClassificationUsing2020a} & ${0.857}$ & ${0.906}$ & ${0.856}$ & $\underline{0.930}(10K)$ \\

			\hline 
			\multirow{4}{*}{ BERT } & 100  &{ISCV}$_{BERT}$ & 0.883 & 0.989 & 0.878 & $0.899(10K)$ \\
			&300& {ISCV}$_{BERT}$ & 0.895 & 0.989 & 0.894 & $0.914(10K)$ \\
			&500& {ISCV}$_{BERT}$ & 0.900 & 0.990 & 0.902 & $0.923(10K)$ \\
			&1000& {ISCV}$_{BERT}$ & $\mathbf{0.907}$ & $\mathbf{0.991}$ & $\mathbf{0.912}$ & $\mathbf{0.937}(10K)$ \\
			\hline
		\end{tabular}
	}
	
	\caption{{The results for full fine-tuning/few-shot approaches. The ``BERT" model, which refers to the bert-base-uncased model. The best result of LOTClass method for each dataset is highlighted by \underline{underlining} the corresponding values. The size of the test set used for the Amazon dataset is denoted by the number in parentheses, where ``$K$" represents a thousand (1000). The notation $\text{ISCV}_{BERT}$ is employed to denote the few-shot-based ISCV method utilizing the bert-base-uncased model. The method name LOTClass with an asterisk (*) indicates that its results were obtained experimentally. The best result for each dataset is indicated by highlighting it in \textbf{bold}.}}
	\label{comparison_result_of_non_prompt_fewshot}
\end{table*}
\subsubsection{{Few-Shot Experiments}}
{
Few-shot based methods are not the primary focus of the ISCV approach. The notable advantage of a well-constructed verbalizer primarily benefits zero-shot based prompt-tuning methods, rather than few-shot or fully fine-tuning based methods \citep{huKnowledgeablePrompttuningIncorporating2022}. In order to showcase the effectiveness of ISCV in the context of few-shot learning using the bert-base-uncased model, we conducted a comparison between few-shot experiments employing ISCV and fully fine-tuned experiments utilizing LOTClass.
}

{
The test set of the Amazon dataset utilized in these experiments consists of 10,000 examples randomly selected from its original test set, as outlined in Table \ref{datasetdetail}. The results in the first row of Table\ref{comparison_result_of_non_prompt_fewshot} are cited from the paper of LOTClass \citep{mengTextClassificationUsing2020a}. In Table \ref{comparison_result_of_non_prompt_fewshot}, the method name "LOTClass*" in the second row indicates the reproduced results obtained using the source code of LOTClass\footnote{https://github.com/yumeng5/LOTClass}. The LOTClass method underwent fine-tuning by utilizing all samples from the training set of each respective dataset. Conversely, the $\text{ISCV}_{BERT}$ method randomly selects a specific number of training examples per category, referred to as "shots," from the train set of each dataset to fine-tune the bert-base-uncased model.
}

{
Table \ref{comparison_result_of_non_prompt_fewshot} showcases that $\text{ISCV}_{BERT}$ demonstrates superior performance by achieving improved results with a mere 100 examples per category across datasets such as AG's News, DBPedia, and IMDB. Similarly, for the Amazon dataset, only 1000 examples are sufficient to attain enhanced outcomes.  With 1000 examples per category, $\text{ISCV}_{BERT}$ surpasses LOTClass by 4.3\%, 8.0\%, 4.7\%, and 0.3\% on datasets AG's News, DBPedia, IMDB, and Amazon, respectively.
This evidence substantiates that the prompt-tuning-based method $\text{ISCV}_{BERT}$ can yield equivalent or superior results with a smaller number of examples. Moreover, it underscores the efficacy of a well-constructed verbalizer in prompt-tuning approaches based on few-shot learning.
}

\section{Discussion}


\subsection{ {Out of Vocabulary (OOV) }}

Our proposed ISCV utilizes calibrated concepts as label words. Essentially, most of the chosen concepts are phrases that are not part of a pre-trained language model's (PLM) vocabulary, and are therefore considered out-of-vocabulary (OOV). Each OOV phrase consists of multiple tokens from the vocabulary. In a masked language model such as BERT \citep{devlinBERTPretrainingDeep2019a} or RoBERTa \citep{liuRoBERTaRobustlyOptimized2019a}, it returns the probability of each token in the vocabulary. Previous studies have employed three strategies for an OOV word: assigning the first value of all tokens, the maximum value of all tokens, or the mean probability value of all tokens in a label word. The mean value is a suitable approach for multi-token phrases, as suggested by \citep{toshniwalCrossTaskAnalysisText2020}. To overcome the OOV issue, we directly calculate the average probabilities of each token within a concept.

\subsection{ {Setting  of Hyperparameters}}

ISCV encompasses four main hyperparameters, including the size of the support set $\mathcal{T}_n$ for named entity extraction (denoted as $n$), the size of the support set $\mathcal{T}_v$ for cascade calibration (denoted as $q$), the top-$j$ number of concepts selected after language model calibration (denoted as $j$), and the number of concepts selected as label words for each class (denoted as $l$). All of these hyperparameters are determined through pilot experiments or experience.

Regarding the hyperparameter $n$, it determines the final number of named entities (or Part-of-Speech tags for sentiment analysis), which are collectively referred to as named entities for simplicity. Through pilot experiments, considering the dataset distribution, we discovered that the optimal range for $n$ falls between 200 and 4000. As a pre-setting value, we have set it to 4000.

The hyperparameter $q$ is set within the range of 100 to 1500. Through pilot experiments, we have discovered that using fewer than 2000 examples can effectively capture the distribution of the dataset being used.

Through reading papers such as \citep{huKnowledgeablePrompttuningIncorporating2022} and testing code related to verbalizer construction, we have determined that a valid verbalizer typically requires fewer than 500 label words per class. Therefore, we have set the range for the hyperparameter $l$ from 10 to 700. By examining the final values of ISCV in Table \ref{tab:hyperparameter_set}, it is evident that an appropriate value for $l$ is less than 500 on the used datasets.

Regarding the hyperparameter $j$, it is used to enhance the efficiency of the category calibration process. To ensure a consistent number of $l$ label words for each class in a dataset with $k$ classes, we set $j$ as $l \times k \times \text{coefficient}$. Empirically, a coefficient value of 10 has been found to be appropriate. It is important to note that a larger value of $j$ does not compromise the effectiveness of the cascade calibration but may slow down the procedure. For ISCV, the hyperparameter $j$ is set as 10,000. The top-$j$ concepts obtained from the language model calibration are then utilized in the category calibration step.

The final step involves determining the optimal $q\text{-}l$ pair for a given dataset. The hyperparameter $q$ is varied within the range of 100 to 1500, with an interval of 100. Similarly, the hyperparameter $l$ is varied within the range of 50 to 700, with an interval of 50. Additionally, the value 10 is added to the candidate set of hyperparameter $l$, resulting in a range from 10 to 700. ISCV utilizes each pairwise combination of $q$ and $l$ to search for the appropriate $q\text{-}l$ pair that yield the best performance for a specific dataset. This search is conducted using zero-shot text classification while keeping a fixed random seed throughout the procedure.

\section{Conclusion}




This study proposed a novel approach for constructing verbalizers in prompt-based text classification tasks by leveraging external concepts as label-word candidates and employing a cascade calibration procedure to optimize them. Our approach addressed the limitations of current methods, which rely solely on the class name and lack multiple perspectives and higher levels of abstraction. 

The experimental results substantiate that our proposed approach exhibits a remarkable superiority over the state-of-the-art methods in zero-shot based topic classification and achieves superior outcomes in sentiment analysis. This showcases a more effective and dependable technique for verbalizer construction.

Despite the significant contributions of this study, there are still opportunities for future research. Firstly, the parameter search process, while effective, can be challenging and time-consuming for researchers. Therefore, exploring ways to automate and improve this procedure could be a promising area of investigation. Secondly, it is worth noting that this study was limited to an English concept database, and further investigations on other languages, such as Chinese, could provide valuable insights.

\bibliography{iscv_arxiv}
	
\end{document}